\documentclass{article}
\usepackage[utf8]{inputenc}
\usepackage{algorithm}
\usepackage{algorithmic}
\usepackage{amsmath}
\usepackage{amssymb}
\usepackage{caption}
\usepackage{graphicx}
\usepackage{tabu}
\usepackage{wrapfig}
\usepackage[numbers]{natbib}

\usepackage[export]{adjustbox}

\DeclareMathOperator*{\argmin}{arg\,min}

\usepackage[final]{nips_2017}

\usepackage[utf8]{inputenc} 
\usepackage[T1]{fontenc}    
\usepackage{hyperref}       
\usepackage{url}            
\usepackage{booktabs}       
\usepackage{amsfonts}       
\usepackage{nicefrac}       
\usepackage{microtype}      

\title{Deep Learning-Guided Image Reconstruction from Incomplete Data}

\author{
  Brendan Kelly \\
  Department of Computer Science\\
  Washington University in Saint Louis\\
  St. Louis, MO 63130 \\
  \texttt{bmkelly@wustl.edu} \\
  \And
  Thomas P. Matthews \\
  Department of Biomedical Engineering \\
  Washington University in Saint Louis \\
  St. Louis, MO 63130 \\
  \texttt{thomas.matthews@wustl.edu} \\
  \And
  Mark A. Anastasio \\
  Department of Biomedical Engineering \\
  Washington University in Saint Louis \\
  St. Louis, MO 63130 \\
  \texttt{anastasio@wustl.edu} \\
}

\begin{document}

\maketitle

\begin{abstract}

An approach to incorporate deep learning within an iterative image reconstruction framework to reconstruct images from severely incomplete measurement data is presented.  Specifically, we utilize a convolutional neural network (CNN) as a quasi-projection operator within a least squares minimization procedure.  The CNN is trained to encode high level information about the class of images being imaged; this information is utilized to mitigate artifacts in intermediate images produced by use of an iterative method. 
The structure of the method was inspired by the proximal gradient descent method, where the proximal operator is replaced by a deep CNN and the gradient descent step is generalized by use of a linear reconstruction operator.    It is demonstrated that this approach improves image quality for several cases of limited-view image reconstruction and that using a CNN in an iterative method increases performance compared to conventional image reconstruction approaches.  We test our method on several limited-view image reconstruction problems. Qualitative and quantitative results demonstrate state-of-the-art performance. 
\end{abstract}

\section{Introduction}

Deep learning (DL) methods have had a profound impact on computer vision and image analysis applications.  Such methods are now ubiquitous and routinely employed for tasks that include image classification \cite{DBLP:journals/corr/HeZRS15, szegedy2016inception}, segmentation \cite{chen2016deeplab, badrinarayanan2015segnet}, and image completion \cite{yeh2016semantic, IizukaSIGGRAPH2017}, to name a few.  Until very recently, however, DL methods have not been utilized for the upstream task of improving the quality of the images themselves that are produced by computed imaging systems.   The process of forming an image from a collection of measurements by use of a computational procedure is referred to as \emph{image reconstruction}.  Almost all modern imaging systems that are utilized in scientific or medical applications are computed in nature and utilize an image reconstruction method to form an image.
 The next wave in DL methodologies for imaging applications is likely to address the important task of improving the quality
 of images produced by computed imaging systems \cite{wang2016perspective}.

In many imaging applications, in order to reduce data-acquisition times or radiation doses in the case of X-ray-based systems, it is desirable to reconstruct an accurate image from an incomplete set of measurements.   Here, an incomplete set of measurements refers to one that does not uniquely specify an image that can produce the measured data, even in the idealized situation where noise or other measurement errors are absent. In such cases, the measurement process is not described by an injective operator and therefore the inverse operator is not defined.
Accordingly, image reconstruction from incomplete data corresponds  to an ill-posed  inverse problem and a regularized solution must be computed.  
Bayesian theory provides a natural framework for computing regularized solutions via penalized least squares estimators by incorporating \emph{a priori} information regarding the sought-after image.
For example, image priors that describe sparseness properties are routinely employed and have proven to be successful for certain classes of problems. They are particularly effective when the measurement model satisfies the mathematical conditions prescribed by compressive sampling theory \cite{CPA:CPA20124}.  However, most computed imaging systems do not satisfy these conditions.
Although recent advances in generative neural networks are exciting and encouraging \cite{nguyen2016plug} \cite{DBLP:journals/corr/RadfordMC15}, the task of specifying an image prior that comprehensively describes the statistical properties of a specified ensemble of images remains a challenging task.
 Human observers can often reliably identify artifacts in reconstructed images.
Codifying this knowledge in a mathematically tractable way such that it can be incorporated into a reconstruction method has proven difficult. 
   These issues have limited the effectiveness of regularized methods for reconstructing images from   inconsistent and/or incomplete measurement data.


DL methods hold great promise for circumventing these limitations and may enable great improvements in image reconstruction performance when the measurement data are incomplete.  As mentioned above, for this problem, there exists a large number of images that can produce the measured data; one of these images is the sought-after `true' image, while the others will contain artifacts and distortions.
The fundamental problem is that it is difficult to design a regularization strategy (i.e., optimization program) that results in computation of the true solution, or some accurate approximation of it.
To circumvent this, DL methods can be naturally employed to learn domain knowledge that is subsequently exploited by the reconstruction method to improve the accuracy of the reconstructed image.
A DL-assisted image reconstruction method would be `intelligent' in the sense that it is able to utilize this domain knowledge to guide it to arrive at a solution that contains greatly reduced artifact levels as compared to a solution computed by use of a traditional reconstruction method. Intuitively, the DL-assisted reconstruction method would be able to avoid producing an image that contains strong artifacts because the DL model knows what those images look like.
As described below, in terms of linear operator theory, the DL model would permit the reconstruction method to accurately estimate the null space component of the image that is invisible to the imaging system.

In this paper, a DL-assisted image reconstruction method for use with incomplete data is proposed and investigated.
Consider that two ensembles of images are available: one ensemble corresponds to artifact free (`true') images while the second contains images that are degraded by artifacts and distortions that are attributable to the incompleteness of the measurement data. A convolutional neural network (CNN) is  trained by use of these data to learn a mapping that relates the true and artifact-laden images.  
This mapping is subsequently embedded within an iterative image reconstruction method to guide the iterative method toward a solution that contains greatly reduced artifact levels.
The proposed approach holds conceptual similarities to the proximal gradient descent method, which alternates between taking a step along the gradient of some cost function and applying a proximal operator. In the proposed approach, the proximal operator is replaced by the mapping learned by the CNN and the gradient descent update is generalized by use of a reconstruction operator.
A novel two-stage training scheme is introduced in order to train the CNN to be used within the proposed framework.

\textbf{Related Work}

Extensive work has been performed on applying DL methods to image restoration tasks, such as denoising or in-painting \cite{Kim_2016_VDSR, yeh2016semantic, NIPS2016_6172}. These tasks share similarities with image reconstruction tasks, but differ in the information provided as input. In image restoration, a degraded image is provided as input, while in image reconstruction, measurements corresponding to some operator applied to the image are provided. Typically, these measurements differ structurally in some significant way from the images themselves. Additionally, many image reconstruction methods require an initial guess of the true image, which itself may be a degraded image.

Image restoration methods can be combined with image reconstruction methods as a post-processing step to try to correct any artifacts in the reconstructed image. However, since these restoration methods accept only an image as input, it is not possible to consider the measured data during this post-processing step. Thus, in summarizing the related work, the methods have been partitioned into those that apply DL-based image restoration methods to independently reconstructed images and those that consider the measured data as part of a larger DL-based image reconstruction approach. To contrast with the proposed approach, the image restoration methods will be referred to as single-pass approaches, as the images pass through DL-based network a single time.



\textit{\textbf{Deep Image Restoration}}: The problem of image restoration using DL techniques has been well-studied, and, as such, it is not possible to mention all the meaningful contributions.  In \cite{Kim_2016_VDSR}, a deep residual network is proposed for the task of image super-resolution.  In \cite{NIPS2016_6172}, an autoencoder style CNN is proposed which makes use of residual-style layers \cite{DBLP:journals/corr/HeZRS15}, and is applied to the tasks of image denoising and super-resolution.  Recently, several groups has have used DL techniques to remove artifacts and noise from images reconstructed from incomplete and/or inconsistent measurement data \cite{chen2017low, antholzer2017deep, wang2016accelerating}.

\textit{\textbf{Deep Image Reconstruction}}: Recently, several groups have proposed using DL techniques to improve image reconstruction.  In \cite{Hershey2014arXiv09}, inference algorithms are unfolded into a series of layers and the parameters of those layers are optimized by a neural network.  This technique has been expanded to incorporate many standard inference algorithms and applications \cite{NIPS2016_6406, perdios2017learning, Hammernik2017, kamilov_learning_2016}.  In \cite{kulkarni2016reconnet}, the original image is directly estimated by a CNN.  In \cite{bora2017compressed}, the original image is estimated by gradient descent where the estimated image is constrained to lie in the range of a generative CNN.  In \cite{meinhardt2017learning}, the proximal operator is replaced by a denoising CNN for an image reconstruction problem.


There has been a large amount of work conducted on understanding and improving the optimization of conventional CNNs for the task of image restoration.  However, many of the deep image reconstruction techniques are more difficult
to train than CNNs, which has limited the complexity of the DL models they employ.  In this work, a DL-assisted reconstruction method is proposed that embeds a conventional CNN within an iterative framework. In this way, the method is able to
exploit the power of a deep CNN that is relatively easy to train but also is responsive to the measured data.

\section{Statement of image reconstruction problem}


Consider an inverse problems in which one seeks to recover an image $\mathbf{f} \in \mathbb{R}^n$ from a collection of measurements $\mathbf{g} \in \mathbb{R}^m $ for $m < n$. While the proposed approach is more general, for ease of discussion and interpretation, we will focus on the case in which the measured data are related to the true image via
\begin{equation}
\label{eq:imaging_model}
\mathbf{g} = \mathbf{H}\mathbf{f},
\end{equation}
where $\mathbf{H} \in \mathbb{R}^{m \times n}$ is a linear operator describing the measurement system. 
Given some $\mathbf{H}$, any object $\mathbf{f}$ can be decomposed into two orthogonal components, the measurable component $\mathbf{f}_{meas} \in \mathbb{U}_{meas}$, and the null space component $\mathbf{f}_{null} \in \mathbb{U}_{null}$, where $\mathbb{U}_{null} \equiv \{ \mathbf{f} ~|~ \mathbf{H}\mathbf{f}  = 0 \} $ and $\mathbb{U}_{meas}$ is its orthogonal complement. A key challenge is how to recover the component of $\mathbf{f}$ that lies in $\mathbb{U}_{null}$.  From the definition of $\mathbb{U}_{null}$, it follows that the measured data $\mathbf{g}$ does not contain any information on $\mathbf{f}_{null}$, which is
 invisible to the imaging system. Thus, to accurately recover $\mathbf{f}_{null}$ and thereby avoid
 strong artifacts in the reconstructed $\mathbf{f}=\mathbf{f}_{meas}+\mathbf{f}_{null}$,
 some \textit{a priori} knowledge about the set of objects that might be measured is needed.

The penalized least squares (PLS) approach to image reconstruction seeks to compute an estimate $\hat{\mathbf{f}}$
of the image by
minimizing an objective function consisting of a data fidelity (loss) term and a penalty, or regularization, term. For example,
\begin{equation}
\hat{\mathbf{f}} = \argmin_{\mathbf{f} \in \mathcal{S}} \| \mathbf{H}\mathbf{f} - \mathbf{g} \|^2_2 + \lambda \Phi(\mathbf{f}),
\label{eqn:pls_problem}
\end{equation}
where $\mathcal{S}$ is some, typically convex, set, $\Phi$ is a regularization term, and $\lambda > 0$ is a scalar parameter that controls the relative weight of the two terms in the objective function. 

\section{Approach}

\textbf{DL-Guided Image Reconstruction}

The structure of the proposed approach is inspired by that of proximal (or projected) gradient descent. In that approach, an initial guess for the true object $\mathbf{f}^{(0)}$ is iteratively refined by first taking a step along the gradient of some cost function $\mathcal{C}\left(\mathbf{f}\right)$ and then applying a proximal operator $\mathcal{P}$ that maps the current iterate to:
\begin{equation}
	\hat{\mathbf{f}}^{(k+1)} = \mathcal{P}\left(\hat{\mathbf{f}}^{(k)} - \alpha \nabla \mathcal{C}\left(\hat{\mathbf{f}}^{(k)}\right) \right) 
\end{equation}
Here, $k$ is the iteration number and $\alpha$ is the step size. In theory, this approach can be very effective. For example, 
$\mathcal{P}$ could be defined as a projection operator onto the set of all artifact-free images. In practice, however, it is difficult to determine a method for computing $\mathcal{P}$ except for relatively simple cases. This limits the amount and complexity of \textit{a priori} information that can be encoded in the projection operator.

To move beyond this restriction, $\mathcal{P}$ is replaced with a quasi-projection operator $\mathcal{Q}\left(\mathbf{f}; \mathbf{w}\right)$, consisting of a CNN parameterized by a set of weights $\mathbf{w}$. These weights will be determined by training the CNN to map images from the space of artifact-laden images to the space of artifact-free images. In this case, the learned operator $\mathcal{Q}$ may not strictly satisfy the properties of a projection or proximal operator. However, as demonstrated below, embedding this quasi-projection operator within an iterative reconstruction
method  can result in images that possess significantly reduced artifact levels as compared to those produced by use conventional (non-DL-based) reconstruction methods.


\begin{wrapfigure}{R}{0.5\textwidth}
	\begin{minipage}{0.5\textwidth}
	    \vspace{-10pt}
		\begin{algorithm}[H]
        	\caption{General reconstruction procedure}
        	\label{alg:general_recon}
        	\begin{algorithmic}[1]
        		\REQUIRE $\mathbf{w}, \mathbf{p}, \mathbf{g}, \mathbf{H}, n, \mathcal{R}, \mathcal{Q}$
        		\ENSURE $\hat{\mathbf{f}}$
        		\STATE{$\hat{\mathbf{f}}_\mathcal{Q}^{(0)} \gets \mathbf{0}$}
        		\STATE{$k \gets 0$} \COMMENT{$k$ is the algorithm iteration number.}
        		\WHILE {$k < n$}
        		\STATE{$\hat{\mathbf{f}}_{\mathcal{R}}^{(k+1)} \gets \mathcal{R}\left( \hat{\mathbf{f}}_\mathcal{Q}^{(k)}; \mathbf{H}, \mathbf{g}, \mathbf{p}\right)$}
        		\STATE{$\hat{\mathbf{f}}_\mathcal{Q}^{(k+1)} \gets \mathcal{Q}
        			\left(\hat{\mathbf{f}}_R^{(k+1)}; \mathbf{w}\right)$}
        		\STATE{$k \gets k + 1$}
        		\ENDWHILE
        		\STATE{$\hat{\mathbf{f}} \gets \hat{\mathbf{f}}_\mathcal{Q}^{(k)}$}
        	\end{algorithmic}
        \end{algorithm}
        \vspace{-10pt}
    \end{minipage}
\end{wrapfigure}

Traditionally, projection or proximal operators are applied at every iteration during the reconstruction process. However, that would require the quasi-projection operator $\mathcal{Q}$ to project the estimated images at early iterations, which may be very inaccurate. Instead, a further generalization of the proximal gradient descent method is proposed. The single gradient descent step performed at each iteration is replaced by a more general approximate reconstruction operator $\mathcal{R}\left(\mathbf{f}; \mathbf{H}, \mathbf{g}, \mathbf{p}\right)$, where $\mathbf{p}$ is a set of parameters for the reconstruction method. This approximate reconstruction operator should update the current estimate of the object to more closely match the measured data. For example, $\mathcal{R}$ could be a solution to the optimization problem given in Eqn.~\ref{eqn:pls_problem} subject to some stopping criteria. 
As a result, the output of $\mathcal{R}$ should be closer to the sought-after image, yielding a simpler quasi-projection mapping for the CNN to learn. In this way, the training problem is made easier, while $\mathcal{Q}$ still learns all the necessary \textit{a priori} information to improve the reconstructed image given by our conventional reconstruction operator $\mathcal{R}$.  Pseudocode for our algorithm is provided in Alg. \ref{alg:general_recon}, where $n$ is the number of times $\mathcal{R}$ and $\mathcal{Q}$ are applied.

When $\mathcal{Q}$ is chosen to be a projection operator and $\mathcal{R}$ is chosen to correspond to a single gradient descent step, the proposed approach reduces to the projected gradient descent method as a special case. Further, when $n = 1$, the proposed approach reduces to the single-pass approach.

\textbf{Network Architecture for Establishing $\mathcal{Q}$} 


The  operator $\mathcal{Q}$ represents a mapping from $\mathbb{R}^n$ to $\mathbb{R}^n$. When employing a DL model to implement this mapping, there is great freedom in selecting the appropriate network architecture.  We base our selection on two main reasons.  The first is that recent work has shown that deeper networks are more powerful than shallower networks with the same number of parameters \cite{DBLP:journals/corr/HeZRS15},\cite{DBLP:journals/corr/SrivastavaGS15}.  Deeper networks consist of a larger chain of non-linear layers which allows for a higher level of abstraction.  The second reason is that learning a residual mapping is an attractive alternative when there is a high correlation between the input and output.  Since the input and output are quite similar for this problem, it is intuitively an easier problem to learn the residual instead of learning the direct mapping.

The chosen network architecture for implementing $Q$ corresponds to a deep residual CNN, inspired by \cite{Kim_2016_VDSR}.  The network contains 20 convolutional layers, with a ReLU layer after each convolutional layer except for the final layer.  In all but the final layer, there are $64$ filters of size $3\times3$.  In the final layer the CNN is mapping back to the original image space so it contains a single $3\times3$ layer. 
Instead of learning a direct mapping, this network predicts a residual image that is added to the original input to get the final output.  Additional details can be found in \cite{Kim_2016_VDSR}.





\textbf{Training $\mathcal{Q}$}

\textit{\textbf{Conventional training:}}  Given a training set of images $\mathcal{F}$ and their corresponding measured data 
$\left\lbrace \mathbf{g} = \mathbf{H}\mathbf{f} ~|~ 
\mathbf{f} \in \mathcal{F} \right\rbrace$, the weights $\mathbf{w}$ are determined by minimizing

\begin{equation}
    \label{eq:stage_1}
    \hat{\mathbf{w}} = \argmin_{\mathbf{w}} \frac{1}{2} \sum_{i} \|\mathbf{f}_{(i)} -  \mathcal{Q}\left(\hat{\mathbf{f}}_{\mathcal{R},(i)}^{(1)}; \mathbf{w}\right)\|_2^2 ,
\end{equation}

where the subscript $(i)$ denotes the $i$-th training sample and $\hat{\mathbf{f}}_{R,(i)}^{(k)}$ denotes the estimate of the  image given by 
line 4 of Alg.~\ref{alg:general_recon} for the $k$-th iteration and 
the $i$-th training sample.  This corresponds to optimizing the weights of the CNN using reconstructed images generated by $\mathcal{R}$ for an initial guess of all zeros.  


\textit{\textbf{Two stage training:}} It was observed that a CNN optimized by the conventional training scheme described above may
not generalize well to 
inputs of $\hat{\mathbf{f}}_{\mathcal{R},(i)}^{(k)}$ for $k > 1$.  In order to regularize the weights, a two-stage training scheme was introduced, where the network is initially trained with only zero-initialized reconstructed images as in Eqn.~\ref{eq:stage_1}.  Then, the weights are fine-tuned with intermediate results acquired by applying Alg.~\ref{alg:general_recon} with the weights trained in the first stage,
\begin{equation}
    \label{eq:stage_2}
    \hat{\mathbf{w}}^* = \argmin_{\hat{\mathbf{w}}} \frac{1}{2} \sum_{i,k} \|\mathbf{f}_{(i)} -  \mathcal{Q}\left(\hat{\mathbf{f}}_{\mathcal{R},(i)}^{(k)}; \hat{\mathbf{w}}\right)\|_2^2 ,
\end{equation}
where $k$ denotes the iteration number during which the reconstructed image was estimated.  During training, no additional improvement is made after including $10$ intermediate results per image, so $n=10$ during this second training stage.  This phenomena is likely due to the second stage acting as a data augmentation technique which produces very similar, and therefore redundant, results after $n=10$ iterations.  Use of dropout or traditional data augmentation techniques lead to worse improvement compared to the two stage training scheme.


A separate CNN was trained for each of the studies described below.  All models were trained in the same way, using ADAM \cite{ADAM}, a batchsize of 64 and minimizing the mean squared error loss as in Eqn.~\ref{eq:stage_1} and \ref{eq:stage_2}.  The default parameters for ADAM \cite{ADAM} were used, except for a learning rate of $0.0001$.  Each model was trained for $3$ million iterations in each stage, taking around 10 hours total on 2 NVIDIA Titan X GPU cards.  We employed the weight initialization strategy introduced in \cite{Xavier}.

\section{Computer-simulation studies}


As a demonstration of our approach, a canonical image reconstruction problem in X-ray computed tomography (CT) that utilizes very limited measurements was considered.

\textbf{Sample generation}

Taking inspiration from the Shepp-Logan phantom \cite{SheppLogan}, the generated images are made up of one main ellipse and between 2 and 7 other minor ellipses. The number and characteristics of the ellipses were chosen randomly. Additional details about the sample generation process are provided in the supplementary materials.  This particular form for the samples allowed use of a closed form solution for calculating the measured data. This provided an independent way to calculate the measured data that did not depend on the numerical model $\mathbf{H}$.    

The images were of size $256 \times 256$, with $7500$ training images, $1000$ validation images, and $500$ testing images.  The validation images were used to optimize hyper-parameters, and the test images where used to evaluate performance only once.

\textbf{X-ray CT forward model}


An idealized 2-D X-ray CT forward model is considered. A series of parallel X-rays are transmitted through a 2-D object with the intensity of the X-rays decreasing as they travel through the object. The resulting intensity of the transmitted X-rays is recorded by a linear detector. The detector is rotated about the object and this process is repeated for a collection of tomographic views.  Typically, the detector is swept across an angular range of 180$^\circ$. Here, we consider the limited-view case in which the angular coverage is much less than 180$^\circ$. 

There exists two main sources of error when reconstructing images.  The first source is model error, where the assumed forward operator, $\mathbf{H}$, does not correspond to the forward operator that generated the measured data.  In practice, $\mathbf{H}$ cannot exactly capture all properties of a real-world imaging system so there is always some amount of model error.  In simulations, we can construct a scenario with no model error by using the same $\mathbf{H}$ in the generation of the measured data by Eqn. \ref{eq:imaging_model} and in the reconstruction.  This scenario is referred to as inverse crime.  The second source of error is noise in the measured data.  This occurs due to imperfections in the measurement equipment, and we can simulate this phenomena by 
treating the measured data as a random vector.

In this work, we simulated three different scenarios: (1) the inverse crime scenario, where we have no model error and no noise in the measured data; (2) a scenario with model error but no stochastic noise; and (3) a scenario that contains both model error and stochastic noise in the measured data.  We simulated noise in the measured data by introducing Gaussian noise to $\mathbf{g}$ with a standard deviation equal to $2\%$ of the maximum of $\mathbf{g}$.

For the non-inverse-crime scenarios, we use three different ranges of tomographic view angles: 60, 100, and 140 degrees, with one view at each degree.  These angular ranges all correspond to a severely under-determined problems that cannot be accurately solved with existing methods.  For the inverse crime case, we only consider the 60 degree case.  In all cases, the detector has 256 elements. 

\vspace{2em}
\textbf{Image reconstruction methods} 

In this work, three different formulations of the optimization problem given by Eqn.~\ref{eqn:pls_problem} were considered for purposes of comparison, corresponding to different choices of $\mathcal{S}$ and $\Phi\left(\mathbf{f}\right)$:
\begin{enumerate}
	\item Least squares (LS): $\mathcal{S} = \mathbb{R}^n$, $\Phi\left(\mathbf{f}\right) = 0$
	\item Non-negative least squares (LS-NN): $\mathcal{S} = \mathbb{R}^n_{\ge 0}$, $\Phi\left(\mathbf{f}\right) = 0$
	\item Penalized least squares with TV regularization (PLS-TV): $\mathcal{S} = \mathbb{R}^n_{\ge 0}$, $\Phi\left(\mathbf{f}\right) = \|\mathbf{f}\|_{TV}$,
\end{enumerate}
where $\mathbb{R}^n_{\ge 0}$ denotes the set of non-negative real numbers and $\|\mathbf{f}\|_{TV}$ denotes the total-variation (TV) semi-norm. 

The proposed DL-assisted approach requires some choice for the approximate reconstruction operator $\mathcal{R}$
(line 4 in Alg.\ 1). In this work,  the action of $\mathcal{R}$ was computed by approximately minimizing the LS objective function or the LS-NN objective function. 

The LS objective function was employed in the inverse crime case. 
The least squares solution was computed by applying the Moore-Penrose pseudoinverse of $\mathbf{H}$, denoted $\mathbf{H}_{MP}^{-1}$, to the measured data. 
The Moore-Penrose pseudoinverse was computed by first performing singular value decomposition of $\mathbf{H}$ and then employing the method described in \cite{barrett2004foundations}. 

The LS-NN objective function was employed in the case of inconsistent data. Empirically, it was observed that this produced superior results in the case of inconsistent data, perhaps as the non-negativity constraint results in estimated images that are closer to the true images. As a result, the mapping that must be learned by the quasi-projection operator $Q$ may be simpler and thus easier to approximate with a given finite-capacity network architecture. In this case, the LS-NN optimization problem was solved by projected gradient descent.


For all iterative methods, a constant step size of 0.75 was employed. The stopping criteria was set to be when the relative change in the $\ell_2$-norm of the object between consecutive iterations was less than 0.001. 

\vspace{-.5em}
\section{Numerical experiments}
\vspace{-.5em}

\textbf{Interpretation}

\begin{wrapfigure}{R}{0.5\textwidth}
    \vspace{-10pt}
	\centering
	\includegraphics[width=.5\textwidth]{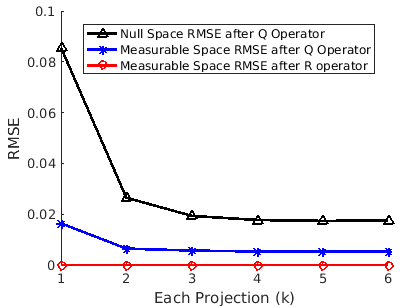}
	\caption{Comparison of RMSE in measurable and null space for a test sample from the 60 degree inverse crime case.  In black, we show the RMSE between the null space component of the intermediate reconstruction $\mathbf{f}_\mathcal{Q}^{k}$ and the true null space component $\mathbf{f}_{null}$.  We show the RMSE between the measurable space components of $\mathbf{f}_\mathcal{Q}^k$ in blue and $\mathbf{f}_{\mathcal{R}}^k$ in red, and the true measurable component $\mathbf{f}_{meas}$.    }
	\label{fig:interpretation_meas_null_IC_60D}
	\vspace{-10pt}
\end{wrapfigure}

The least challenging version of the inverse problem, the inverse crime case, was first considered.  
The   operator $\mathcal{R}$ was specified to compute the solution of the LS optimization problem. This solution can be calculated by applying the Moore-Penrose pseudoinverse to the measured data. 
Since the measured data is consistent and $\mathbf{H}_{MP}^{-1} \mathbf{H}$ is the projection operator onto $\mathbb{U}_{meas}$, $\mathcal{R}$, in this case, recovers the measurable component of the true object. The null space component of the estimated object will remain unchanged from that of the initial guess.  Further, any alterations to measurable component made by $\mathcal{Q}$ can be corrected by applying $\mathcal{R}$ while leaving the null component unchanged. In this way, $\mathcal{R}$ and $\mathcal{Q}$ each take responsibility for estimating different components of the object: the measurable component is estimated by $\mathcal{R}$ and the null component is estimated by $\mathcal{Q}$. By applying each of these operators in an alternating fashion, the information in the measured data and the learned \textit{a priori} information encoded in the network are jointly considered, allowing both the measurable space and null space components to be more accurately reconstructed. 

This behavior is reflected in the plots of the RMSEs in the measurable and null space components, shown in Fig.~\ref{fig:interpretation_meas_null_IC_60D}. The RMSE of the null space component is progressively improved with repeated application of $\mathcal{Q}$. The RMSE of the measurable space component estimated by $\mathcal{R}$ (e.g. before application of $\mathcal{Q}$) is quite small for all iterations, as this component can be reliably estimated from the measured data. The RMSE of the measurable space component is increased after applying the $\mathcal{Q}$ operator, but repeated application of $\mathcal{R}$ recovers the high accuracy estimate of the measurable space component. With iteration, the method converges to an estimate that has low RMSE in both the measurable and null space components.



When $\mathcal{R}$ does not correspond to the solution of the LS optimization problem, $\mathcal{R}$ may make changes to the estimated null space component. Still, the division of responsibility in which $\mathcal{R}$ estimates the measurable space component and $\mathcal{Q}$ estimates the null space component may largely, though perhaps not strictly, hold true. While this approach does not necessarily have as simple an interpretation, it may be beneficial for two reasons. First, if the output of $\mathcal{R}$ is closer to the true object, the mapping that must be learned by the $\mathcal{Q}$ operator may be simpler, resulting in an easier training task. Second, the learned network may not strictly enforce simple constraints, such as non-negativity, which can be readily included within a conventional reconstruction approach. By incorporating these constraints within $\mathcal{R}$, both learned and user-specified \textit{a priori} information can be included within the proposed framework.


\textbf{Results}

The proposed approach was compared with a single-pass approach and PLS-TV method.  In the DL-assisted approach,  little to no increase in performance was observed after $5$ iterations; thus $n=5$ was employed. 

\begin{figure}[h]
    \centering
	\includegraphics[width=\textwidth,trim={0 9.8cm 0 0},clip]{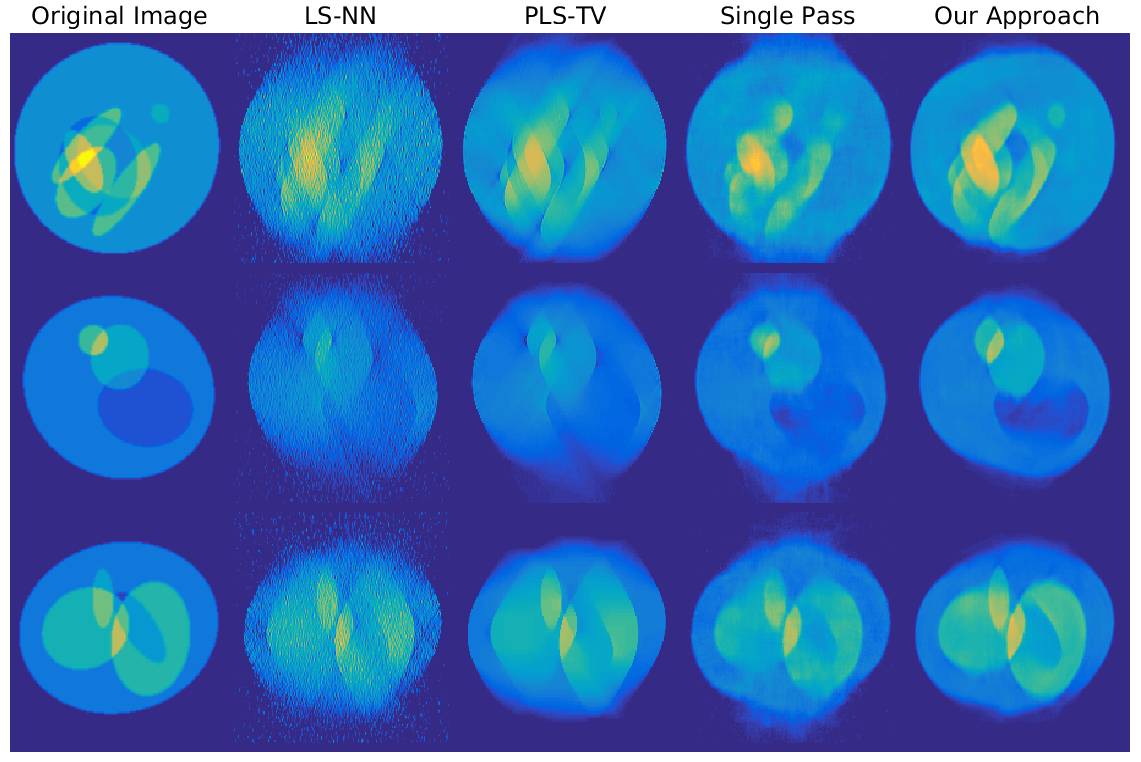} 
	\caption{Images showing showing reconstruction results for all methods for several test images for the 60 degree noisy case.  The images are shown in false color, with a consistent intensity scale.}
	\label{fig:qual_results_noiseless}
\end{figure}

In order to compare with the single-pass approach (i.e., CNN-based image restoration), a CNN 
with the same architecture described above was initially trained using the conventional training scheme.  The performance of the single-pass CNN increased when trained with the two stage scheme.  This performance increase can be explained by the additional training data acting in the capacity of common data augmentation techniques which increase the ability of the CNN to generalize to new data.  Therefore, to fairly compare the single-pass approach with the proposed DL-assisted reconstruction approach, the same CNN is used for both methods.

\begin{figure}[h]
    \centering
	\includegraphics[width=\textwidth,trim={0 9.8cm 0 0},clip]{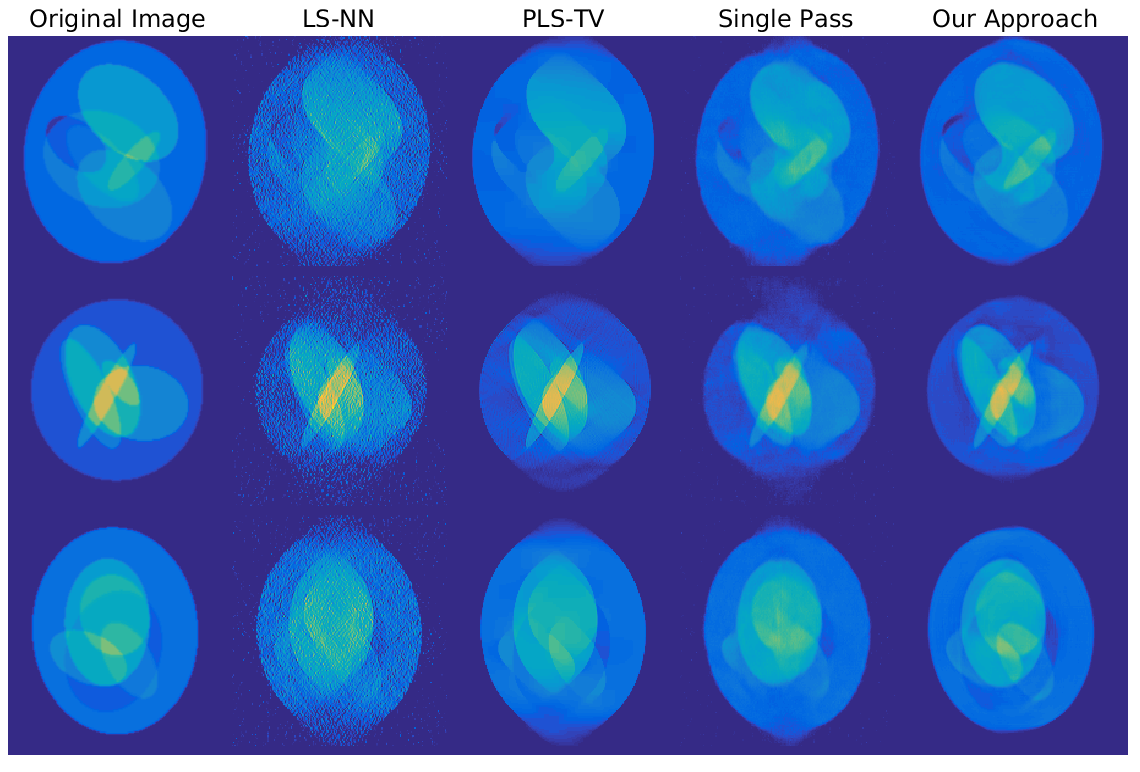} 
	\caption{Images showing showing reconstruction results for all methods for several test images for the 100 degree noisy case.  The images are shown in false color, with a consistent intensity scale.}
	\label{fig:qual_results_noiseless2}
	\vspace{-1em}
\end{figure}

The PLS-TV optimization problem was solved using the FISTA \cite{doi:10.1137/080716542} with an initial guess of all zeros. This approach represents a state-of-the-art method for sparse image reconstruction \cite{sidky2008image}. It should be especially effective for the piece-wise constant images considered in this work, which possess relatively sparse gradient maps. The regularization parameter was swept using a grid search and the image corresponding to the lowest RMSE was selected for comparison with the proposed approach.

Representative reconstructed images for the 60 and 100 degree noisy cases are displayed in Figs.~\ref{fig:qual_results_noiseless} and \ref{fig:qual_results_noiseless2}. None of the example images considered were contained in the training set.  Additionally,  results for the other cases are presented in the supplementary material.  The PLS-TV reconstruction was quite well suited to remove noise from the measured data, but it was not as successful at removing streaking artifacts.  It is apparent that a single-pass with the CNN significantly improved the accuracy of the reconstructed images, but the proposed approach provided substantial further improvement.  


\begin{table}[h]
	\label{sample-table}
	\centering
	\begin{tabular}{lllllllll}
		\toprule
		~ & \multicolumn{4}{c}{RMSE} & \multicolumn{4}{c}{SSIM}                 \\
		\cmidrule(l){2-5} \cmidrule(l){6-9}
		~     & LS & PLS-TV & SP & Proposed & LS & PLS-TV & SP & Proposed \\
		\midrule
		60D IC & 0.111 & 0.097 & 0.046 & \textbf{0.027} & 0.490 & 0.772 & 0.823 & \textbf{0.929}     \\
		\midrule
		~      & LS-NN   & PLS-TV    & SP & Proposed & LS-NN & PLS-TV    & SP & Proposed \\
		\midrule
		60D     & 0.112 & 0.099 & 0.072 & \textbf{0.057} & 0.689 & 0.734 & 0.796 & \textbf{0.874}       \\
		100D    & 0.069 & 0.059 & 0.031 & \textbf{0.022} & 0.764 & 0.852 & 0.929 & \textbf{0.969}  \\
		140D    & 0.050 & 0.042 & 0.017 & \textbf{0.011} & 0.793 & 0.895 & 0.963 & \textbf{0.989} \\
		\midrule
		60D N   & 0.154 & 0.010 & 0.089 & \textbf{0.074} & 0.261 & 0.734 & 0.665 & \textbf{0.794} \\
		100D N  & 0.151 & 0.061 & 0.057 & \textbf{0.047} & 0.277 & 0.845 & 0.758 & \textbf{0.887} \\
		140D N & 0.155 & 0.049 & 0.032 & \textbf{0.026} & 0.296 & 0.894 & 0.810 & \textbf{0.900}\\
		\midrule
	\end{tabular}
	\caption{Average RMSE/SSIM for each reconstruction approach for test set.  IC - Inverse Crime. N - Noise applied to measured data. D - Degree, the number of view. SP - Single-pass approach.}
	\label{table:results_table}
	\vspace{-2em}
\end{table}

For each case, the performance was evaluated on 500 test images.  The average root mean squared error (RMSE)
and structural similarity (SSIM) were computed as summary measures of performance.  RMSE corresponds to the pixel-wise difference between the original image and the reconstructed image, 
while SSIM attempts to measure differences between images in a manner more consistent with human visual perception.
As indicated in Table \ref{table:results_table}, the proposed approach produces the best measures for all test cases.  
Interestingly, the performance of PLS-TV varied little between the noisy and noiseless cases, which suggests it performed extremely well at removing the noise from the measured data.  The single-pass and the proposed approach performed worse in the noisy cases compared with the noiseless cases, with the single-pass approach achieving worse SSIM than PLS-TV for the  noisy cases.

The LS-NN reconstructions are significantly worse than PLS-TV reconstructions, yet the DL approaches performed quite well using the LS-NN as the $\mathcal{R}$ operator.  This suggests a significant possible performance increase in using PLS-TV in the DL approaches, as PLS-TV successfully removes stochastic noise from the measured data.


\vspace{-1em}
\section{Conclusion}
\vspace{-1em}

An approach was presented that embeds a conventional deep CNN within an iterative framework for the task of image reconstruction.  Due to its iterative nature, our approach balances the information available in the measured data with the learned \textit{a priori} information encoded within the CNN.  Experimental results and our analysis show our approach achieves better results than a common single-pass CNN-based image restoration approach and a state-of-the-art reconstruction method that exploits sparsity for regularization.
Our framework is general enough to incorporate other variants of $\mathcal{R}$, such as ones that
estimate a PLS-TV solution, and can be applied to fields outside image reconstruction as well.


\begin{thebibliography}{10}

\bibitem{antholzer2017deep}
Stephan Antholzer, Markus Haltmeier, and Johannes Schwab.
\newblock Deep learning for photoacoustic tomography from sparse data.
\newblock {\em arXiv preprint arXiv:1704.04587}, 2017.

\bibitem{badrinarayanan2015segnet}
Vijay Badrinarayanan, Alex Kendall, and Roberto Cipolla.
\newblock Segnet: A deep convolutional encoder-decoder architecture for image
  segmentation.
\newblock {\em IEEE Transactions on Pattern Analysis and Machine Intelligence},
  2017.

\bibitem{barrett2004foundations}
H.H. Barrett and K.J. Myers.
\newblock {\em Foundations of image science}.
\newblock Wiley series in pure and applied optics. Wiley-Interscience, 2004.

\bibitem{doi:10.1137/080716542}
Amir Beck and Marc Teboulle.
\newblock A fast iterative shrinkage-thresholding algorithm for linear inverse
  problems.
\newblock {\em SIAM Journal on Imaging Sciences}, 2(1):183--202, 2009.

\bibitem{bora2017compressed}
Ashish Bora, Ajil Jalal, Eric Price, and Alexandros~G Dimakis.
\newblock Compressed sensing using generative models.
\newblock {\em arXiv preprint arXiv:1703.03208}, 2017.

\bibitem{CPA:CPA20124}
Emmanuel~J. Candès, Justin~K. Romberg, and Terence Tao.
\newblock Stable signal recovery from incomplete and inaccurate measurements.
\newblock {\em Communications on Pure and Applied Mathematics},
  59(8):1207--1223, 2006.

\bibitem{chen2017low}
Hu~Chen, Yi~Zhang, Mannudeep~K Kalra, Feng Lin, Peixi Liao, Jiliu Zhou, and
  Ge~Wang.
\newblock Low-dose ct with a residual encoder-decoder convolutional neural
  network (red-cnn).
\newblock {\em arXiv preprint arXiv:1702.00288}, 2017.

\bibitem{chen2016deeplab}
Liang-Chieh Chen, George Papandreou, Iasonas Kokkinos, Kevin Murphy, and Alan~L
  Yuille.
\newblock Deeplab: Semantic image segmentation with deep convolutional nets,
  atrous convolution, and fully connected crfs.
\newblock {\em arXiv preprint arXiv:1606.00915}, 2016.

\bibitem{Xavier}
Xavier Glorot and Yoshua Bengio.
\newblock Understanding the difficulty of training deep feedforward neural
  networks.
\newblock {\em International conference on artificial intelligence and
  statistics}, 2010.

\bibitem{Hammernik2017}
Kerstin Hammernik, Tobias W{\"u}rfl, Thomas Pock, and Andreas Maier.
\newblock {\em A Deep Learning Architecture for Limited-Angle Computed
  Tomography Reconstruction}, pages 92--97.
\newblock Springer Berlin Heidelberg, Berlin, Heidelberg, 2017.

\bibitem{DBLP:journals/corr/HeZRS15}
Kaiming He, Xiangyu Zhang, Shaoqing Ren, and Jian Sun.
\newblock Deep residual learning for image recognition.
\newblock In {\em The IEEE Conference on Computer Vision and Pattern
  Recognition (CVPR)}, June 2016.

\bibitem{Hershey2014arXiv09}
John~R. Hershey, Jonathan Le~Roux, and Felix Weninger.
\newblock Deep unfolding: Model-based inspiration of novel deep architectures.
\newblock September 2014.
\newblock arXiv:1409.2574.

\bibitem{IizukaSIGGRAPH2017}
Satoshi Iizuka, Edgar Simo-Serra, and Hiroshi Ishikawa.
\newblock {Globally and Locally Consistent Image Completion}.
\newblock {\em ACM Transactions on Graphics (Proc. of SIGGRAPH 2017)},
  36(4):107:1--107:14, 2017.

\bibitem{SheppLogan}
Anil~K. Jain.
\newblock {\em Fundamentals of Digital Image Processing}, page 439.
\newblock Englewood Cliffs, NJ, Prentice, 1989.

\bibitem{kamilov_learning_2016}
U.~S. Kamilov and H.~Mansour.
\newblock Learning {Optimal} {Nonlinearities} for {Iterative} {Thresholding}
  {Algorithms}.
\newblock {\em IEEE Signal Processing Letters}, 23(5):747--751, May 2016.

\bibitem{Kim_2016_VDSR}
Jiwon Kim, Jung~Kwon Lee, and Kyoung~Mu Lee.
\newblock Accurate image super-resolution using very deep convolutional
  networks.
\newblock In {\em The IEEE Conference on Computer Vision and Pattern
  Recognition (CVPR Oral)}, June 2016.

\bibitem{ADAM}
Diederik~P. Kingma and Jimmy Ba.
\newblock Adam: {A} method for stochastic optimization.
\newblock {\em CoRR}, abs/1412.6980, 2014.

\bibitem{kulkarni2016reconnet}
Kuldeep Kulkarni, Suhas Lohit, Pavan Turaga, Ronan Kerviche, and Amit Ashok.
\newblock Reconnet: Non-iterative reconstruction of images from compressively
  sensed random measurements.
\newblock {\em arXiv preprint arXiv:1601.06892}, 2016.

\bibitem{NIPS2016_6172}
Xiaojiao Mao, Chunhua Shen, and Yu-Bin Yang.
\newblock Image restoration using very deep convolutional encoder-decoder
  networks with symmetric skip connections.
\newblock In D.~D. Lee, M.~Sugiyama, U.~V. Luxburg, I.~Guyon, and R.~Garnett,
  editors, {\em Advances in Neural Information Processing Systems 29}, pages
  2802--2810. 2016.

\bibitem{meinhardt2017learning}
Tim Meinhardt, Michael M{\"o}ller, Caner Hazirbas, and Daniel Cremers.
\newblock Learning proximal operators: Using denoising networks for
  regularizing inverse imaging problems.
\newblock {\em arXiv preprint arXiv:1704.03488}, 2017.

\bibitem{nguyen2016plug}
Anh Nguyen, Jason Yosinski, Yoshua Bengio, Alexey Dosovitskiy, and Jeff Clune.
\newblock Plug \& play generative networks: Conditional iterative generation of
  images in latent space.
\newblock {\em arXiv preprint arXiv:1612.00005}, 2016.

\bibitem{perdios2017learning}
Dimitris Perdios, Adrien Georges~Jean Besson, Philippe Rossinelli, and
  Jean-Philippe Thiran.
\newblock Learning the weight matrix for sparsity averaging in compressive
  imaging.
\newblock In {\em IEEE International Conference on Image Processing (ICIP
  2017)}, number EPFL-CONF-225655, 2017.

\bibitem{DBLP:journals/corr/RadfordMC15}
Alec Radford, Luke Metz, and Soumith Chintala.
\newblock Unsupervised representation learning with deep convolutional
  generative adversarial networks.
\newblock {\em CoRR}, abs/1511.06434, 2015.

\bibitem{sidky2008image}
Emil~Y Sidky and Xiaochuan Pan.
\newblock Image reconstruction in circular cone-beam computed tomography by
  constrained, total-variation minimization.
\newblock {\em Physics in medicine and biology}, 53(17):4777, 2008.

\bibitem{DBLP:journals/corr/SrivastavaGS15}
Rupesh~Kumar Srivastava, Klaus Greff, and J{\"{u}}rgen Schmidhuber.
\newblock Highway networks.
\newblock {\em CoRR}, abs/1505.00387, 2015.

\bibitem{szegedy2016inception}
Christian Szegedy, Sergey Ioffe, Vincent Vanhoucke, and Alex Alemi.
\newblock Inception-v4, inception-resnet and the impact of residual connections
  on learning.
\newblock {\em arXiv preprint arXiv:1602.07261}, 2016.

\bibitem{wang2016perspective}
Ge~Wang.
\newblock A perspective on deep imaging.
\newblock {\em IEEE Access}, 4:8914--8924, 2016.

\bibitem{wang2016accelerating}
Shanshan Wang, Zhenghang Su, Leslie Ying, Xi~Peng, Shun Zhu, Feng Liang, Dagan
  Feng, and Dong Liang.
\newblock Accelerating magnetic resonance imaging via deep learning.
\newblock In {\em Biomedical Imaging (ISBI), 2016 IEEE 13th International
  Symposium on}, pages 514--517. IEEE, 2016.

\bibitem{NIPS2016_6406}
Yan Yang, Jian Sun, Huibin Li, and Zongben Xu.
\newblock Deep admm-net for compressive sensing mri.
\newblock In D.~D. Lee, M.~Sugiyama, U.~V. Luxburg, I.~Guyon, and R.~Garnett,
  editors, {\em Advances in Neural Information Processing Systems 29}, pages
  10--18. Curran Associates, Inc., 2016.

\bibitem{yeh2016semantic}
Raymond Yeh, Chen Chen, Teck~Yian Lim, Mark Hasegawa-Johnson, and Minh~N Do.
\newblock Semantic image inpainting with perceptual and contextual losses.
\newblock {\em arXiv preprint arXiv:1607.07539}, 2016.

\end{thebibliography}
\end{document}